\documentclass{article}
\usepackage{ijcai20}

\usepackage{times}
\usepackage{soul}
\usepackage{url}
\usepackage[utf8]{inputenc}
\usepackage[small]{caption}
\usepackage{graphicx}
\usepackage{amsmath}
\usepackage{amsthm}
\usepackage{booktabs}
\usepackage{algorithm}
\usepackage{algorithmic}

\usepackage{epsfig}
\usepackage{graphicx}
\usepackage{amssymb}
\usepackage{subfigure}
\usepackage{color}
\usepackage{multirow}
\usepackage{caption} 

\title{SelectScale: Mining More Patterns from Images via Selective and Soft Dropout}

\author{
    Anonymous Submission
    \affiliations
    Paper ID 4356
}

\author{
Zhengsu Chen$^1$\and
Jianwei Niu$^{1,2,3}$\and
Xuefeng Liu$^1$\footnote{Contact Author}\And
Shaojie Tang$^4$\\
\affiliations
$^1$Beihang University\\
$^2$Hangzhou Innovation Institute of Beihang University\\
$^3$Zhengzhou University\\
$^4$University of Texas at Dallas\\
\emails
\{danczs, niujianwei, liu\_xuefeng\}@buaa.edu.com,
tangshaojie@gmail.com
}
\begin{document}

\maketitle

\begin{abstract}
Convolutional neural networks (CNNs) have achieved remarkable success on image recognition. Although the internal patterns of the input images are effectively learned by the CNNs, these patterns only constitute a small proportion of useful patterns contained in the input images. This can be attributed to the fact that the CNNs will stop learning if the learned patterns are enough to make a correct classification. Network regularization methods like dropout and SpatialDropout can ease this problem. During training, they randomly drop the features. These dropout methods, in essence, change the patterns learned by the networks, and in turn, forces the networks to learn other patterns to make correct classification. However, the above methods have an important drawback. Randomly dropping features is generally inefficient and can introduce unnecessary noise. To tackle this problem, we propose SelectScale. Instead of randomly dropping units, SelectScale selects the important features in networks and adjusts them during training. Using SelectScale, we improve the performance of CNNs on CIFAR and ImageNet.
\end{abstract}

%\begin{highlights}
%\item Research highlights item 1
%\item Research highlights item 2
%\item Research highlights item 3
%\end{highlights}
%
%\begin{keywords}
%Dropout \sep
%Network regularization \sep
%Convolutional neural networks \sep 
%Image classification\sep 
%Deep learning \sep
%
%\end{keywords}

%
%\maketitle

\section{Introduction}
Recently, deep neural networks have enabled breakthroughs on computer vision and natural language processing tasks~\cite{girshick2014rich,long2015fully,graves2013speech}. With specially designed architecture, convolutional neural networks (CNNs) have been widely used on image classification~\cite{szegedy2015going,he2016deep}, object detection~\cite{ren2015faster} and semantic segmentation~\cite{long2015fully}. Since AlexNet~\cite{krizhevsky2012imagenet} significantly improved the classification performance on ImageNet, many new deep convolutional neural network architectures (VGG~\cite{simonyan2014very}, Inception~\cite{szegedy2015going,szegedy2016rethinking}, ResNet~\cite{he2016deep}, DenseNet~\cite{huang2016densely} ) are proposed. CNNs extract the features of the input data automatically and the classifier makes decisions according to the features.

%In CNNs, the features are extracted automatically and .
%The representations learned by these networks expose the semantic patterns of the input data. These  representations have many excellent invariance properties.
%remarkable success on recognition tasks and usually employ lots of parameters.

\begin{figure}[t!]

\begin{center}

\centering
\includegraphics[width=0.95\linewidth,height=0.98\linewidth]{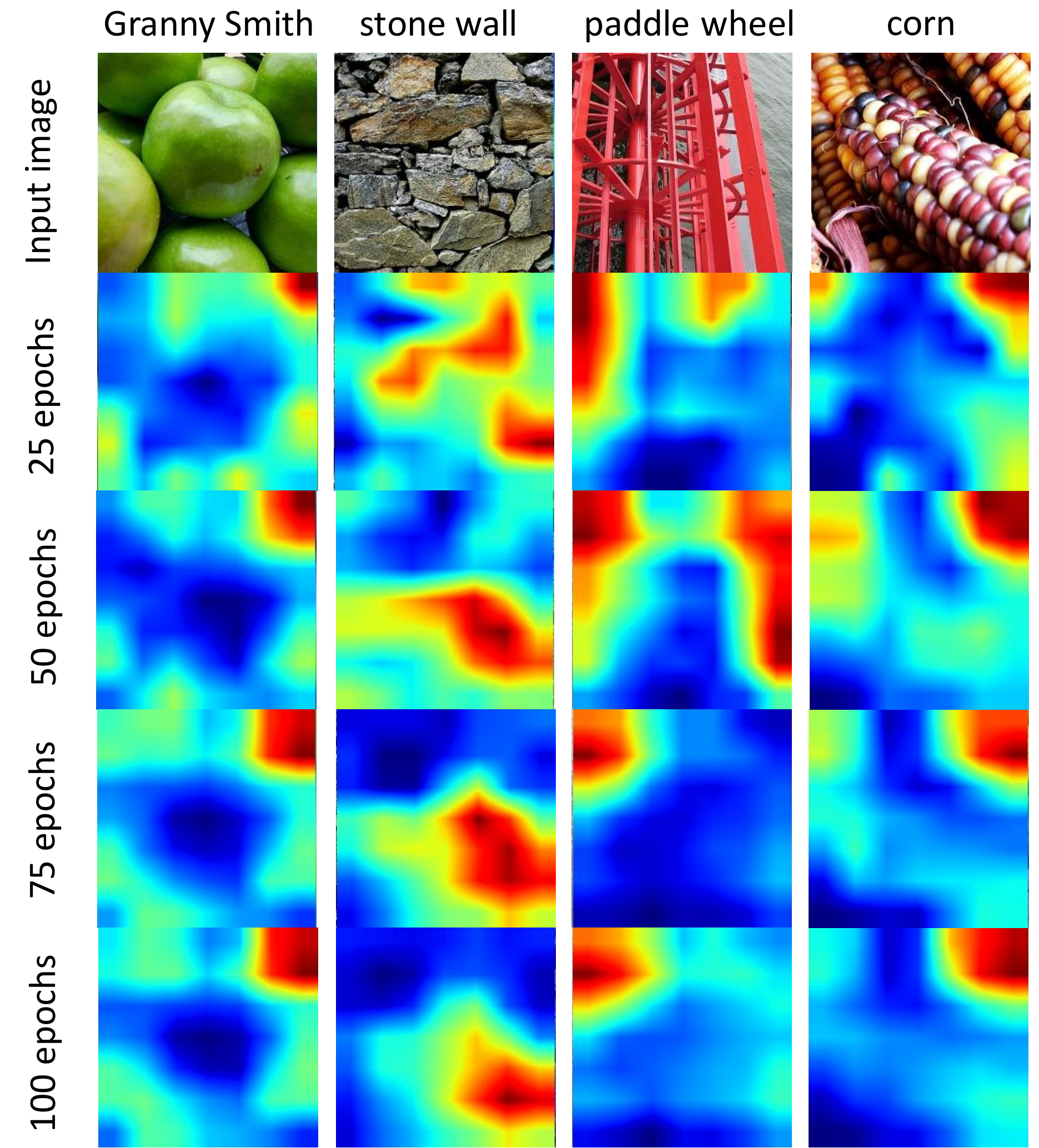} 

\end{center}
\begin{flushleft}
\caption{The class activation maps of training examples for different training epochs. During training, the networks may only utilize a small subset of all useful patterns in images. Even if they grasp rich patterns in the beginning, they can lose some of them during training and overfit to the remaining patterns.}
\label{fig:cam-train}
\end{flushleft}
%\label{fig:cam-train} %% label for entire figure

\end{figure}

\begin{figure*}
\centering
\begin{center}
\label{fig:drop_methods} %% label for entire figure
\subfigure[]{
\label{fig:nodrop:a} %% label for first subfigure
\includegraphics[width=0.3\linewidth]{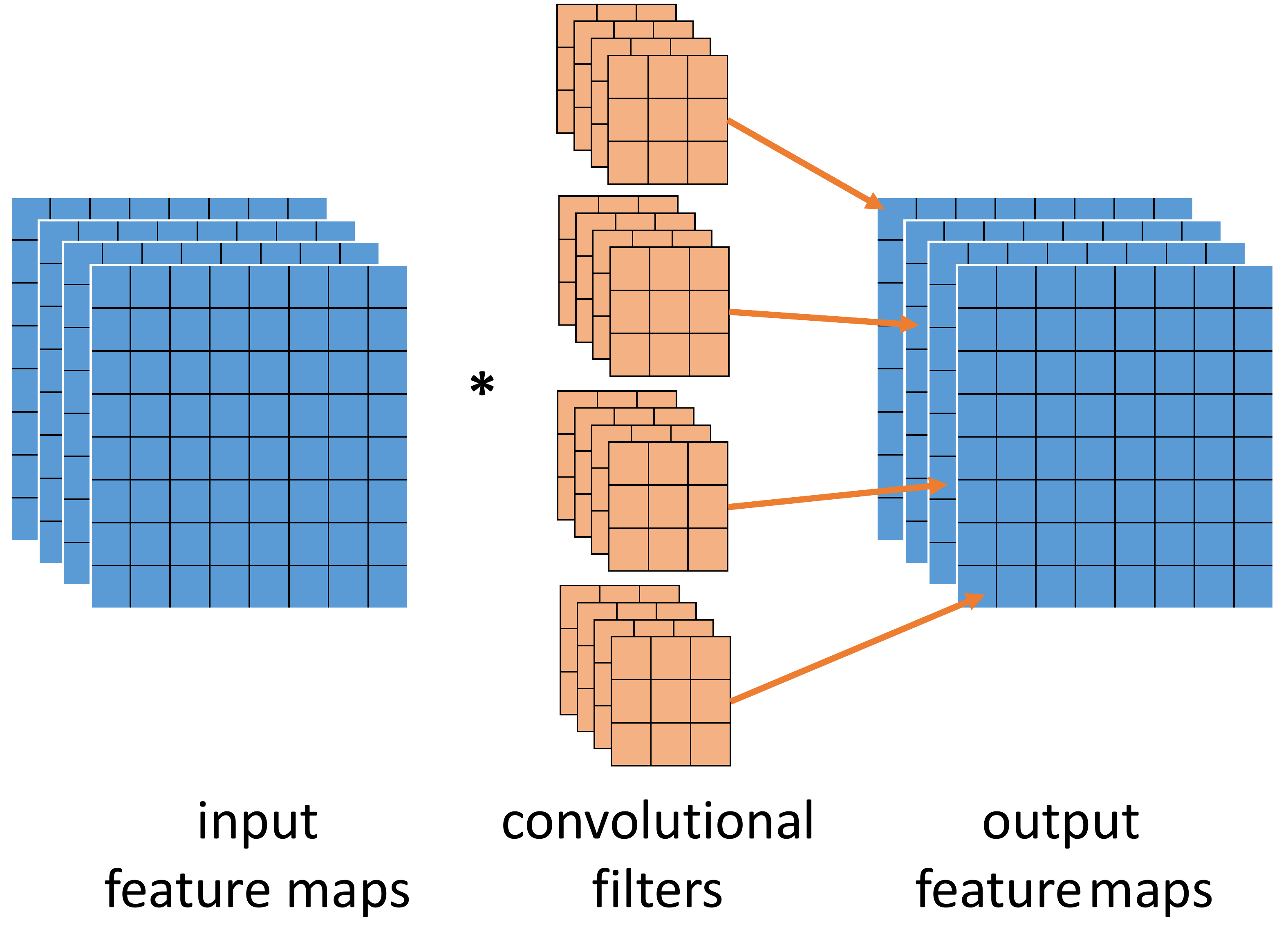}} %matipath1.pdf
\hspace{0.2in}
\subfigure[]{
\label{fig:dropout:b} %% label for second subfigure
\includegraphics[width=0.3\linewidth]{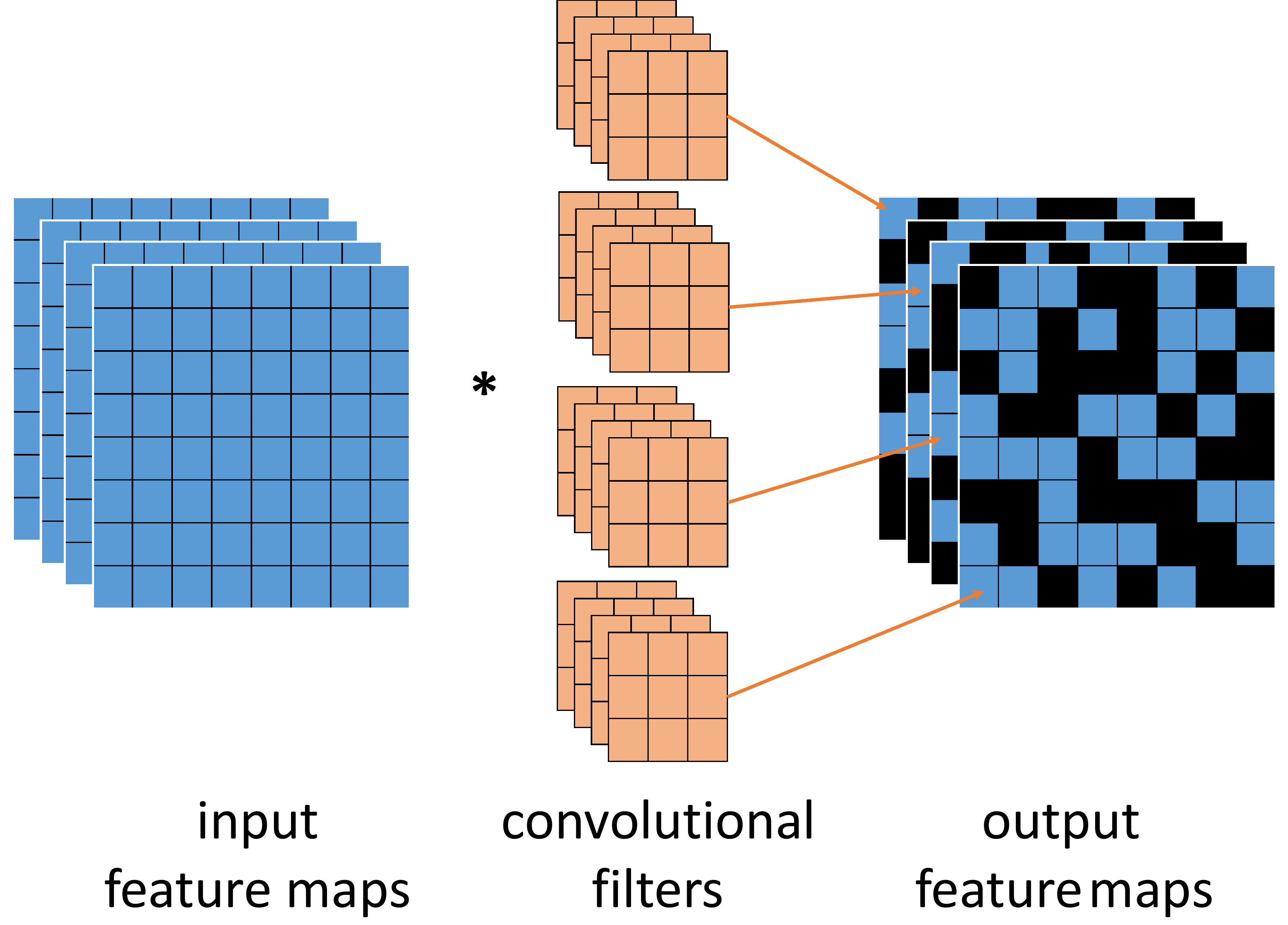}} %matipath2.pdf
\hspace{0.2in}
\subfigure[]{
\label{fig:spatialdropout:c} %% label for second subfigure
\includegraphics[width=0.3\linewidth]{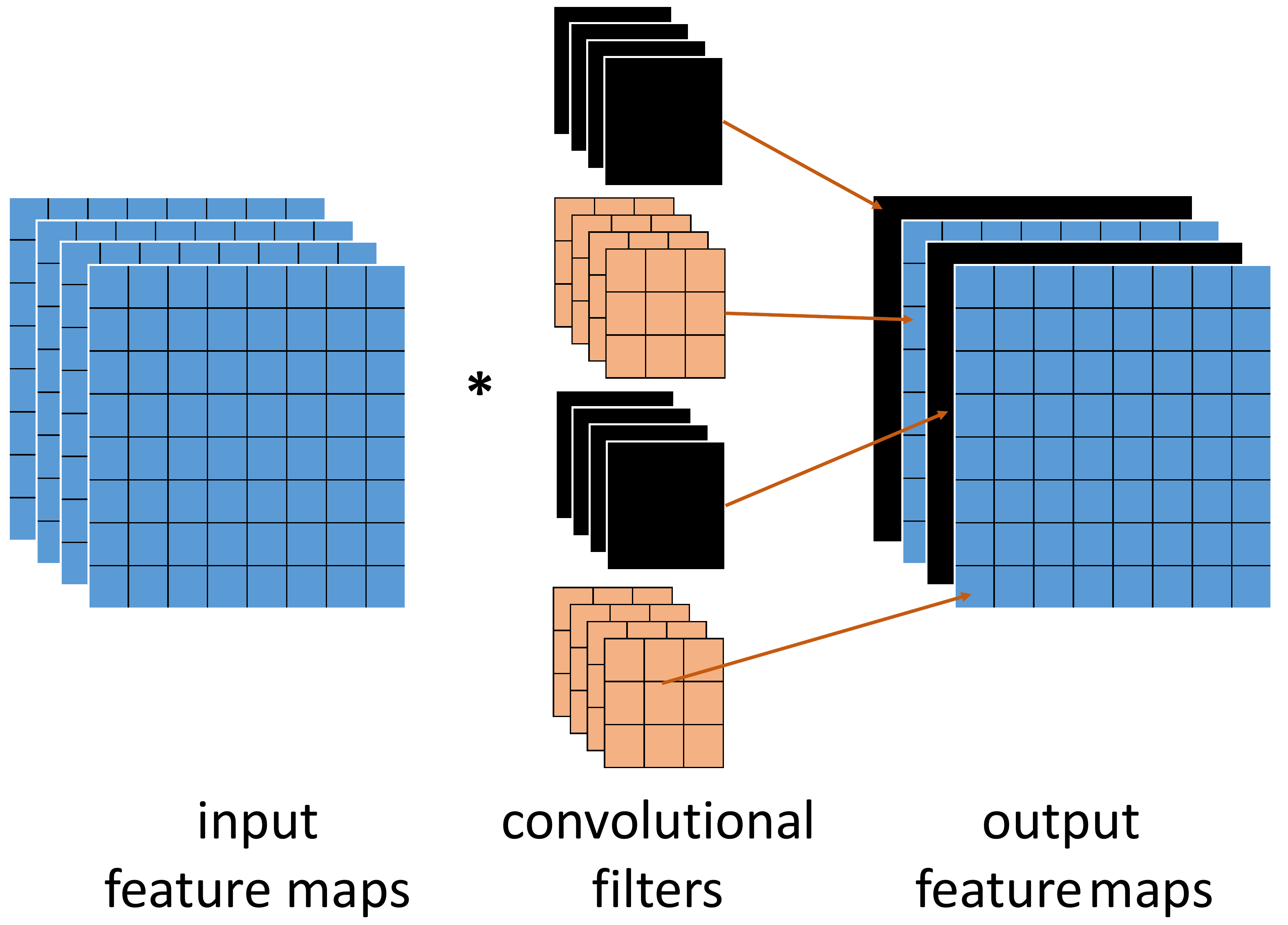}} %matipath2.pdf
\end{center}

\caption{~(a)~A standard $3\times3$ convolutional layer.~(b)~Standard dropout for outputs. The dropped elements are shown in black. The retaining rate is 0.5.~(c)~SpatialDropout for outputs. The dropped feature maps are shown in black. It is equal to dropping the corresponding filters. The retaining rate is 0.5.}

\end{figure*}

\begin{figure}[thbp]

\begin{center}

\centering
\includegraphics[width=0.8\linewidth]{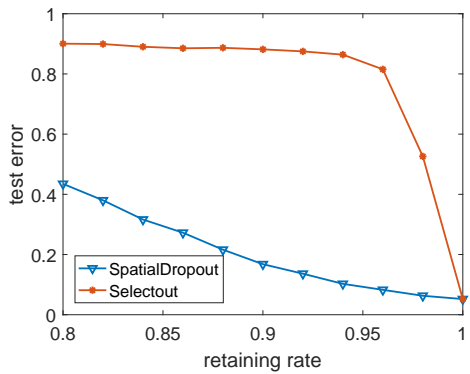} 

\end{center}
\begin{flushleft}
\caption{~The test error for WRN-20-4 with different data drop methods on CFIAR-10 for inference.}
\label{fig:sdrop_cdrop}
\end{flushleft}
%\label{fig:} %% label for entire figure

\end{figure}

However, CNNs usually can not utilize all of the useful patterns in the input data for image classification. If they have learned enough patterns for classification, they will stop learning from the samples. Figure~\ref{fig:cam-train} shows the class activation maps (CAM)~\cite{zhou2016learning} of training images for ResNet-50 on ImageNet. As can be seen, during training the network only learns a subset of patterns from these samples. However, there still exist many patterns yet to be observed by the CNNs. These patterns, if added into the CNNs, will increase the robustness and accuracy of the networks.  

We observed that some regularization techniques can ease the problem above. The most popular one is dropout~\cite{srivastava2014dropout}. Dropout randomly suppresses the outputs of neural units in networks. For a well-learned sample, dropout will change its features. If the learned patterns are destroyed by dropout, the networks need to learn new patterns to recognize the input image. Thus a network with dropout generally outperforms the original networks. 

However, dropout suppresses the output randomly. It may not drop the data as we want. First, it may drop the features of unlearned patterns. Under this circumstance, it has no contribution to learning new patterns but introduces unnecessary noise to the networks. Second, dropout may drop too many important features. The remaining information may be not enough for recognition, which may mislead the networks and make the networks unstable.

To address this problem, we propose Selectout. Selectout suppresses the neural units selectively instead of dropping them randomly. During training, we design a fast and parameter-free method to select the features that are important for recognition. Selectout changes these important features during training, therefore the networks need to learn new useful patterns to classify it. For Selectout, the dropped features are under control. First, it only drops the features that are important for the networks, which avoids introducing unnecessary noise. Second, Selectout can determine how many important features to drop. Thus, the features will not be seriously destroyed by the drop method. Selectout is a more effective data drop strategy. 

Since SpatialDropout~\cite{tompson2015efficient} is more effective than dropout in dropping the features, we construct Selectout with SpatialDropout. SpatialDropout directly drops feature maps instead of dropping neural units. The difference between standard dropout and SpatialDropout is shown in Figure~\ref{fig:dropout:b} and \ref{fig:spatialdropout:c}. 
% SpatialDropout is introduced by~\cite{tompson2015efficient} to regularize a convolutional layer for object localization.
%SpatialDropout directly drops feature maps instead of dropping neural units. In other words, SpatialDropout randomly suppresses some convolution filters of a convolutional layer. In convolutional neural networks, the outputs in the same feature map share the same filter. The filters can be seen as the basic units for feature extraction. Thus SpatialDropout is more effective in dropping the features in convolutional neural networks, which is coincident with what we observe in the experiments.
Based on SpatialDropout, Selectout needs to select the important feature maps in a convolutional layer.

Selectout is a very strong regularization method because it drops the important features in the networks. To roughly show the regularization capacity, we train a network without drop method and apply data drop methods to it during test on CIFAR-10. As shown in Figure~\ref{fig:sdrop_cdrop}, Selectout increases the test error more rapidly than SpatialDropout as the retaining rate decreases. The feature maps selected by Selectout are much more important than the feature maps that are randomly selected by SpatialDropout. Only dropping 6\% of the feature maps by Selectout almost disables the network. In summary, our selection approach for important feature maps is very effective and Selectout is a very strong regularization method.

The regularization of Selectout based on SpatialDropout can be too strong for the networks. The networks are usually very sensitive to the data retaining rate and hard to be tuned. To overcome this problem, we propose SelectScale. SelectScale is based on SpatialScale. Unlike SpatialDropout, SpatialScale scales the feature maps with random weights rather than directly sets some of them to zeros. SpatialScale is similar to SpatialDropout but is milder. It is more stable and easier to be tuned than SpatialDropout. SelectScale is the selecting drop method based on SpatialScale. It only scales the selected feature maps. SelectScale is the main method used in this paper.

SelectScale based on SpatialScale is tested on ResNets, WRNs, ResNeXt and non-residual networks in this paper. It is observed that SelectScale can consistently improve the networks and outperform standard dropout, SpatialDropout, and SpatialScale. 

To summarize, our main contributions are as follows:
\begin{itemize}
\item We propose a novel data drop method that drops the data selectively rather than randomly, which encourages the networks to learn more patterns. 
\item We present a fast and parameter-free method to rank the feature maps during data drop and propose a mild way of dropping data.
\item We conduct detailed experiments to qualitatively and quantitatively demonstrate that our data drop method can significantly improve the network performance.

\end{itemize}

\section{Methods}
In this section, we first review data drop methods and construct Selectout based on SpatialDropout (section~\ref{sec_selectout}). Since Selectout is too strong for many networks, we introduce our mild data drop method: SelectScale (section~\ref{sec_selectscale}). Finally, we detail the function of selecting feature maps when dropping data (section~\ref{subsec:selectingfunction}).

\subsection{Selectout based on SpatialDropout}
\label{sec_selectout}
Given an input image, $\textbf{x}$ is the input of a convolutional layer for a CNN. $c$ is the filter number of this layer. $Y$ are the output feature maps and $W =\{ \textbf{w}_{i} | 1\leq i \leq c\}$ are the weights of the filters in this layer. We have:
\begin{align}
	\textbf{y}_{i} & = f(\textbf{w}_{i}* \textbf{x} + b_{i}) \\
	Y &= \{ \textbf{y}_{i} | 1\leq i \leq c \} \nonumber \label{FilterDrop1}
\end{align}
where $i$ indexes the filters and the output feature maps in the layer. $f$ is the activation function. $b_{i}$ is the bias.
 
%Here, the structure of $\textbf{x}$ is reconstructed according to the filters. Consequently, the convolution can be presented with multiplication.

Using standard dropout, the outputs will be masked by a Bernoulli distribution:
\begin{align}
	\textbf{y}^{dropout}_{i} &= \textbf{r}_{i}\odot \textbf{y}_{i} / p \\
	\textbf{r}_{i} & \sim Bernoulli(p) \nonumber\\ 
	 p       & \in [0,1] \nonumber\\
	 Y_{dropout} &= \{ \textbf{y}^{dropout}_{i} | 1\leq i \leq c\nonumber \}\label{Dropout}
\end{align}
where $p$ is the data retaining rate. The shape of $\textbf{r}_{i}$ is the same as $\textbf{y}_{i}$. To compensate the change of data distribution by dropout, the outputs are scaled by $1/p$ during training~\cite{abadi2016tensorflow}. An alternative method is scaling the outputs by $p$ during testing instead~\cite{srivastava2014dropout}.   

Using SpatialDropout, the outputs: 
\begin{align}
	\textbf{y}^{SpatialDropout}_{i} & =  r_{i}\textbf{y}_{i}/p \\
	r_{i} & \sim Bernoulli(p) \nonumber\\ 
	 p       & \in [0,1] \nonumber\\
	 Y_{SpatialDropout} &= \{ \textbf{y}^{SpatialDropout}_{i} | 1\leq i \leq c \}\nonumber
\label{FilterDrop2}
\end{align}
where $r_{i}$ is the weight to determinate whether to drop the feature map $\textbf{y}_{i}$. Like standard dropout, data after SpatialDropout is also scaled by $1/p$ during training and data drop will be removed during testing.

SpatialDropout regards the convolution filter rather than the neural unit as the basic data drop unit. In convolutional neural networks, the units in the same feature map share the same weights. Thus SpatialDropout directly drops feature maps to regularize the networks as shown in Figure~\ref{fig:spatialdropout:c}. This is the main difference between standard dropout and SpatialDropout.

%Dropout aims to avoid the co-adaptations between units. It randomly suppresses the neural units in a convolutional layer equally.
%SpatialDropout focuses on the co-adaptations between filters. It randomly drops whole feature maps to avoid the co-adaptations between filters. 

%Unlike fully connected layers, convolutional layers have more complex structure. The co-adaptations is also different from that in fully connected layers. We believe that this is the main reason why dropout does not work well for convolutional layers. According to the structure of convolutions, DropFilter is designed to reduce the co-adaptations in convolutional layers. DropFilter regards the filter rather than the neural unit as the basic co-adaptive unit. This is the main difference between dropout and DropFilter.  

Based on SpatialDropout, Selectout is:
%	p &= \frac{1}{c}\sum_{1}^{c}r_{i} \nonumber \\
\begin{align}
	\textbf{y}^{Selectout}_{i} & = \left. r_{i}\textbf{y}_{i} \middle/ p\right. \\
	\textbf{r} &= S( Y,t,p)  \nonumber\\
	{r}_{i} &\in \{0,1\}  \nonumber \\
	 Y_{Selectout} &= \{ \textbf{y}^{Selectout}_{i} | 1\leq i \leq c \}\nonumber\label{selectout}
\end{align}
where $S$ is the selecting function, which determinates how to drop the feature maps. $S$ first generates a score for each feature map and ranks the feature maps with the scores. Then $S$ selects some of the feature maps according to $t$ and $p$. $t$ is the top rate and $p$ is the data retaining rate. For example, using $t = 0.2$ and $p = 0.9$ indicates that $10\%$ of the feature maps are dropped and these dropped feature maps are randomly selected from the top $20\%$ feature maps. $t$ and $p$ are the hyperparameters that can be tuned.
$S$ is detailed in section \ref{subsec:selectingfunction}. 

\subsection{SelectScale based on SpatialScale }
\label{sec_selectscale}
SpatialDropout directly suppresses the outputs of filters. Since SpatialDropout with Selectout will supply stronger regularization, a very small drop rate (e.g. 0.02) can be enough for some networks. However, sometimes the feature maps in convolutional layers can be too few to reveal the difference between very small drop rates. The regularization supplied by Selectout may be unstable and the hyperparameters can be hard to be tuned. We, therefore, propose SpatialScale, which randomly scales the feature maps. Using SpatialScale:
\begin{align}
	\textbf{y}^{SpatialScale}_{i} & =  r_{i}\textbf{y}_{i} \\
	r_{i} & \sim Uniform(1-q,1+q) \nonumber\\ 
	 q       & \in [0,1]\nonumber\\
	  Y_{SpatialScale} &= \{ \textbf{y}^{SpatialScale}_{i} | 1\leq i \leq c \}\nonumber\label{SpatialScale}
\end{align}
The outputs do not need to be scaled, because the mean value of $r_{i}$ is $1$.
During testing, SpatialScale is removed. 

Both SpatialDropout and SpatialScale focus on changing the outputs to regularize the convolutional layers. SpatialDropout directly suppresses the outputs. SpatialScale randomly scales the outputs instead. 
SpatialScale is milder and more stable than SpatialDropout. It changes the output data modestly.

Based on SpatialScale, SelectScale is :
\begin{align}
	\textbf{y}^{SelectScale}_{i} & =  r_{i}\textbf{y}_{i} \\
	r_{i} & =  (1 - v_{i})(u_{i} - 1) + 1 \nonumber\\
	u_{i} & \sim Uniform(1-q,1+q) \nonumber\\ 
	 q       & \in [0,1] \nonumber\\
	 \textbf{v} & = S(Y,t,1-t) \nonumber\\
	 {v}_{i} &\in \{0,1\}  \nonumber \\
	  Y_{SelectScale} &= \{ \textbf{y}^{SelectScale}_{i} | 1\leq i \leq c \}\nonumber\label{SelectScale}
\end{align}
SelectScale only scales the feature maps that are selected by function $S$. The selecting function is detailed in section~\ref{subsec:selectingfunction}.

\subsection{Selecting function $S$}
\label{subsec:selectingfunction}
The selecting function takes the feature maps ($Y$), the top rate ($t$) and the data retaining rate ($p$) as the inputs. In function $S$, we first calculate the absolute values of the feature maps. For the feature maps after ReLU, this step can be removed. Then, the maximum value in each feature map is calculated and these values are regarded as the score of corresponding feature maps. Next, the feature maps are ranked by the scores and the feature maps with top $t$ scores are selected as candidates. For Selectout, the retaining feature maps are randomly selected from the candidates according to the data retaining rate $p$. For SelectScale, all of the candidates will be selected ($p=1-t$). 

%We use $1 - p \leq t$ for all of our experiments. $1 - p >t $ is also tested, for which all the candidates and some other feature maps are selected. It leads to suboptimal results.

In selecting function $S$, the maximum value in each feature map is used as the ranking score. Great activation values usually mean that useful patterns are detected. The results in Figure~\ref{fig:sdrop_cdrop} prove that the feature maps with greater maximum values do play more important roles in neural networks.

Many other methods for obtaining the scores are tested. For example, we apply an average pooling layer before calculating the maximum value. Under these circumstances, the feature maps with greater average local activation values have higher scores. We test the average pooling layer with different pooling windows and strides. Global average pooling is also tested, which utilizes the average activation values of the feature maps as the scores. However, the performance gets worse when enlarging the pooling window and directly adopting the maximum values outperforms these methods.

\begin{figure*}[t]
\label{fig:diffpq}
\begin{center}
\subfigure[]{
\label{fig:diffpq:a} %% label for first subfigure
\includegraphics[width=0.4\linewidth]{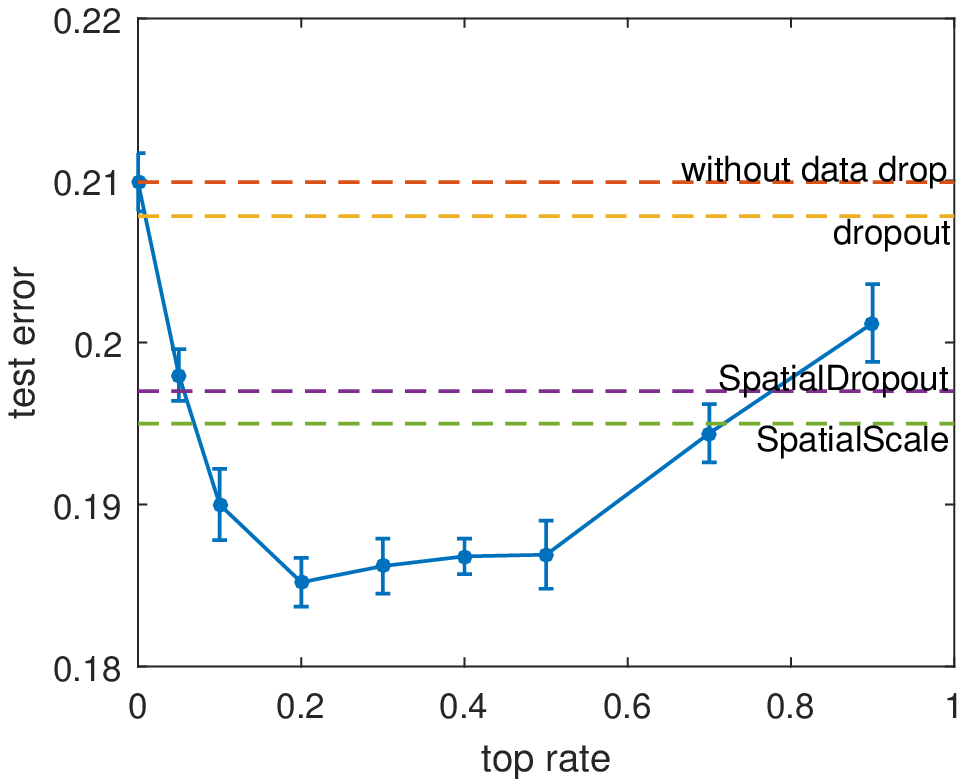}}
\hspace{0in}
\subfigure[]{
\label{fig:diffpq:b} %% label for second subfigure
\includegraphics[width=0.4\linewidth]{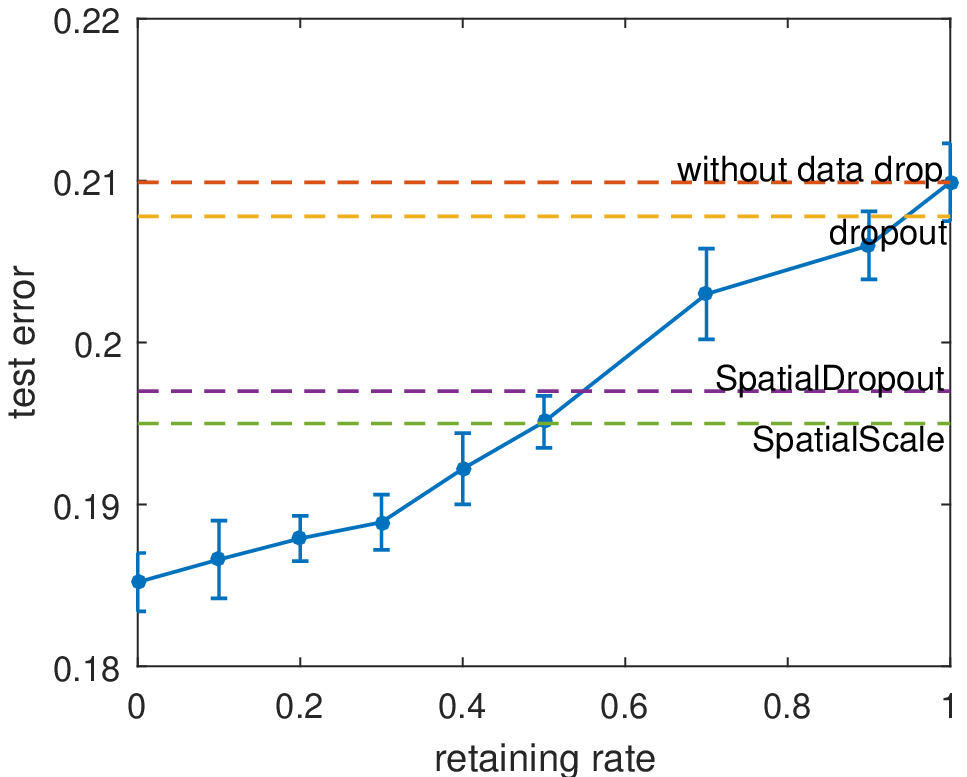}}
\end{center}
\begin{flushright}
\caption{The test errors of different top rates and data retaining rates for WRN-20-8 on CIFAR-100. The best results of other methods are shown.~(a)~The test errors of the network with different top rates and a fixed retaining rate~($q=0$).~(b)~The test errors of the network with different retaining rates and a fixed top rate~($t=0.2$). Note that $retaining rate=0$ in SelectScale indicates that the top-ranked features are scaled by factors sampled from $Uniform(0, 2)$.}
\end{flushright}
\label{fig:path_blocks} %% label for entire figure
\end{figure*}

\section{Experiments}
\subsection{Datasets and settings}
%\subsubsection{Datasets}
CIFAR-10 and CIFAR-100 are colored $32\times32$ natural image datasets that consist of 50,000 images in the training set and 10,000 images in the testing set. Following the common practice~\cite{he2016deep,zagoruyko2016wide}, these tiny images are padded with four pixels on each side and are randomly cropped with a $32\times32$ window afterwards.
% Before being fed into the networks, the images are subtracted by channel means and divided by $128$. 
If not specified, the networks for CIFAR are trained for 200 epochs with 0.1 initial learning rate. The learning rate drops by 0.2 at 60, 120 and 160 epochs following \cite{zagoruyko2016wide}. The batch size is 128 and the weight decay is 0.0005. The data drop method is applied to every convolutional layer for ResNets and WRNs. For ResNeXt, data drop is used before and after the $3\times3$ group convolutions.

%\subsubsection{ImageNet}
ILSVRC2012 is a subset of ImageNet~\cite{russakovsky2015imagenet} database. This dataset contains 1.3M images in the training set and 50K images in the validation set. The top-1 and top-5 error rates on the validation set are reported in this paper. On ImageNet, the standard color augmentation in \cite{russakovsky2015imagenet} is utilized. Following the common practice~\cite{szegedy2016rethinking,he2016identity}, horizontal flip, scale, and aspect ratio augmentation~~\cite{szegedy2016rethinking} is utilized in the experiments. The batch size is 256 and the weight decay is 0.0001. The model is trained for 105 epochs with a 0.1 initial learning rate. The learning rate drops by 0.1 at 30, 60, 90 and 100 epochs. Data drop regularization is adopted for every convolutional layers in bottleneck residual blocks. Like~\cite{szegedy2016rethinking}, the data drop methods are only utilized for the last two groups in ResNet-50.

%%ILSVRC2012 is a quite large dataset and requires many resources to run very deep networks upon it. Like miniImageNet~\cite{vinyals2016matching}, we randomly select 100 classes in ILSVRC2012 and randomly select 600 images in each class. Unlike miniImageNet, the images are not resized. During training the images and their horizontal flips are cropped by $224\times224$ window after the short edge being resized to $256$ following~\cite{simonyan2014very}. The trained models are test on the validation set in which the images that do not belong to the selected 100 classes are removed.

%\captionsetup[table]%{labelformat=simple, labelsep=period}

%\subsection{Implementation details}
%The method in this paper is tested on ResNets~\cite{he2016identity}, WRNs~\cite{zagoruyko2016wide}, and ResNeXt~\cite{xie2016aggregated}. 
We use the template in \cite{he2016deep} to construct the ResNets on CIFAR. 
%The ResNets have three stages. Each stage consists of $n$ residual blocks. Each block contains two $3\times3$ convolutional layers. The filter numbers in the three stages are $\{16,32,64\}$ respectively. The networks with different depths can be constructed by adopting different $n$. The ResNets used in this paper are the pre-activation ResNets, if not specified.
% Like \cite{zagoruyko2016wide}, 
WRNs are constructed by widening the ResNets in this paper. WRN-20-4 represents the network with four times more filters in each layer than ResNet-20.
%The code is available at \url{https://github.com/danczs/SelectScale}.

% All of the networks are implemented by tensorflow~\cite{abadi2016tensorflow} and tensorpack~\cite{wu2016tensorpack}.

%During training the images and their horizontal flips are cropped by $224\times224$ window after the short edge being resized to $256$. For scale augmentation [], the short edge is randomly selected in [256, 480] during training and during testing, the 320x320 crop test results are reported to avoid bias.

\subsection{Different top rate and retaining rate for SelectScale}
To reveal the effects of the two hyperparameters (top rate and retaining rate), SelectScale is tested with different top rates ($t$) and retaining rates ($q$) in this section. The experiments are conducted on CIFAR-100 with WRN-20-8. Fig.~\ref{fig:diffpq:a} shows results of different top rates with a fixed retaining rate ($q=0$). Fig.~\ref{fig:diffpq:a} presents the results of adopting different retaining rates with a fixed top rate ($t=0.2$). The results of the network without data drop and the best results of adopting other drop methods are also shown in the figures. As can be seen, SelectScale outperforms the best results of other methods in a wide range of top rates and retaining rates.

 In Fig.~\ref{fig:diffpq:a}, the network performs best when utilizing a quite small top rate (0.2). It indicates that regularizing the important feature maps is more effective than regularizing all the feature maps. Changing the important feature maps can encourage the network to learn more patterns, while changing the unimportant feature maps may introduce unnecessary noise during training.

\subsection{Results on ResNets and WRNs}
In this section, SelectScale is tested with ResNets and WRNs on CIFAR-10 and CIFAR-100. We apply the drop methods to every convolutional layer. The networks are tested using different retaining rates and the best results are shown in Table~\ref{resnet10} and Table~\ref{resnet100}.

As shown in Table~\ref{resnet10}, SpatialDropout and SpatialScale consistently outperform standard dropout for all networks. When only regularizing the important feature maps by SelectScale, the regularization is more effective and the networks achieve better results. WRN-20-4 with SelectScale achieves a comparable result with WRN-20-8 using SpatialDropout. Note that WRN-20-8 employs about four times more parameters than WRN-20-4. When SelectScale is used for WRN-20-8, its performance is further improved. 

%For ResNet-56, DropFilter gets a poor result because it is too coarse for this network. ScaleFilter achieves a better result with a relatively low $q$ (0.6). 

On CIFAR-100, the improvement of SelectScale is more remarkable. As shown in Table~\ref{resnet100}, standard drop can only provide modest improvement. SpatialDropout and SpatialScale perform much better than standard dropout. SelectScale further reduces the test error by about 1\% for wide residual networks. Note that the top rates for ResNet-110, WRN-20-4 and WRN-20-8 are 0.1, 0.2 and 0.1. In other words, SelectScale regularizes much fewer feature maps than SpatialScale and achieves better results.

\begin{table*}%[width=.95\textwidth]%,pos=h
%\captionsetup[table]{labelsep=quad,margin=10p}
%\captionsetup[table*]{
%  labelsep=quad,
%  justification=justified,
%  labelfont=bf
%}
\caption{Test error of different CNNs on CIFAR-10. The data retaining rate (~$p$~) of standard dropout and SpatialDropout is 0.9 for all tested networks. For SpatialScale, the retaining rate ($q$) is 0.4. The top rates ($t$) and data retaining rates ($q$) of these three networks are 0.4, 0.2, 0.2 and 0.4, 0, 0.1. The results are the average of 5 runs.}

%\caption{Test error of different CNNs on CIFAR-10. The data retaining rate (~$p$~) of standard dropout and SpatialDropout is 0.9 for all tested networks. For SpatialScale, the retaining rate ($q$) is 0.4. The top rate ($t$) and data retaining rate ($q$) of these three networks are 0.4, 0.2, 0.2 and 0.4, 0, 0.1. The results are the average of 5 runs.}
  \label{resnet10}
  \centering
  \begin{tabular}{lccccc}
    \toprule
    Network     & without data drop(\%) & standard dropout(\%) & SpatialDropout(\%)  & SpatialScale(\%) & SelectScale(\%) \\
    %\midrule
    %ResNet-56& 6.20& 6.08 & {\color{blue}6.07} & {\color{red} 5.65} &\\
    \midrule
     ResNet-110& 5.88 $\pm$ 0.13& 5.78 $\pm$ 0.11 & 5.42 $\pm$ 0.15 & {\color{blue} 5.40 $\pm$ 0.06} & {\color{red} 5.22 $\pm$ 0.08} \\
     \midrule
     WRN-20-4& 5.04 $\pm$ 0.07& 4.84 $\pm$ 0.06 & {\color{blue} 4.55 $\pm$ 0.07 } & 4.64 $\pm$ 0.12& {\color{red}4.24 $\pm$ 0.05} \\
     \midrule
     WRN-20-8& 4.93 $\pm$ 0.08& 4.37 $\pm$ 0.07 & 4.21 $\pm$ 0.11 & {\color{blue} 4.14 $\pm$ 0.05} & {\color{red} 3.89 $\pm$ 0.10} \\

    \bottomrule
  \end{tabular}
\end{table*}

\begin{table*}%[width=.95\textwidth]%,pos=h]
  \caption{Test error of different CNNs on CIFAR-100. The data retaining rate (~$p$~) of standard dropout and SpatialDropout is 0.9 for all tested networks. For SpatialScale, the retaining rate ($q$) is 0.4 except for WRN-20-8 ( 0.2 ). The top rates ($t$) and data retaining rates ($q$) of these three networks are 0.1, 0.1, 0.2 and 0.1, 0.1, 0.1. The results are the average of 5 runs.}
  \label{resnet100}
  \centering
  \begin{tabular}{lccccc}
    \toprule
    Network     & without data drop(\%) & standard dropout(\%) & SpatialDropout(\%)  & SpatialScale(\%) & SelectScale(\%) \\
  %  \midrule
  %  ResNet-56&28.07&	27.74&	{\color{blue} 25.83}&	{\color{red} 25.58} & \\
    \midrule
     ResNet-110& 26.16 $\pm$ 0.14&	25.85 $\pm$ 0.24&	24.12 $\pm$ 0.29&	{\color{blue}24.12 $\pm$ 0.19} & {\color{red}23.48 $\pm$ 0.22}\\
     \midrule
     WRN-20-4& 22.90 $\pm$ 0.21 &	22.53 $\pm$ 0.23&	{\color{blue}21.24 $\pm$  0.17}&	21.40 $\pm$ 0.08 & {\color{red}20.20 $\pm$ 0.18} \\
     \midrule
     WRN-20-8& 20.99 $\pm$ 0.18&	20.78 $\pm$ 0.19&	19.70 $\pm$ 0.27&	{\color{blue}19.50 $\pm$ 0.23} & {\color{red}18.52 $\pm$ 0.22} \\

    \bottomrule
  \end{tabular}
\end{table*}

\subsection{SelectScale with other methods}

In this section, more experiments are conducted to demonstrate that SelectScale does not conflict with new residual network architecture and other training methods. The methods utilized in the experiments are followings: 
\begin{itemize}
\item \textbf{New Residual Architecture}.
 ResNeXt is a new residual network architecture that using group convolutions for bottleneck residual blocks~\cite{xie2016aggregated}. It introduces cardinality to residual networks and achieves better results than ResNets upon many visual recognition tasks. 
% ResNeXt-29(8x64d) is used to test our data drop method in this paper. ResNeXt-29(8x64) uses group convolutions for the $3\times3$ convolutions. The group number is 8 and each group consists of 64 filters.
\item \textbf{Cosine Annealing for Learning Rate}.
Cosine annealing for learning rate is introduced by~\cite{loshchilov2016sgdr}. Instead of dropping the learning rate with a factor after each training stage, they drop the learning rate more smoothly by cosine annealing. 
%Afterwards, this method is used in many researches~\cite{gastaldi2017shake,zoph2017learning}.
\item \textbf{Curriculum Dropout}.
Curriculum dropout is introduced by~\cite{Morerio2017Curriculum}. In their opinion, the co-adaptations will not occur at the beginning of training, because the networks are initialized randomly. Thus they linearly decrease the data retaining rate during training. 
%In the experiments in this paper, the dropout and SpacialDropout retaining rate is decreased from 1.0 to 0.6. The top rate of SelectScale is 0.3 and the data retaining rate is decreased from 1.0 to 0.
%We linearly decrease the data retaining rate of standard dropout and DropFilter as well.
\item \textbf{More Training Epochs}. Training the network for more epochs usually will improve the network performance on CIFAR-10~\cite{loshchilov2016sgdr}. The training epochs are increased from 200 to 600 in our experiments.
\end{itemize}
%\item Gradient drop
%Like data drop, gradient drop can also improve the networks sometimes.

The results on CIFAR-10 is shown in Table~\ref{resnext}. As can be seen, because of introducing too much unnecessary noise, the test error is increased by standard dropout. SpatialDropout can improve the networks but SelectScale achieves the best result. 
%DropFilter focuses on reducing the co-adaptations between filters. It does not introducetoo much useless noise and can still improve the networks. 
 
%The results of competitive methods on CIFAR-10 are shown in Table~\ref{state-of-the-art}. As can be seen, improving the results on CIFAR-10 is much difficult. ResNeXt-29 (16x64d) only outperforms ResNeXt-29 (8x64d) by 0.7\%. SelectScale improves ResNeXt-29 (8x64d) nearly without introducing extra computation. Only Shake-Shake-26~(2x96d) outperforms our model. But Shake-Shake-26~(2x96d) needs to be trained for three times more epochs(1800 epochs).

\begin{table}%[width=.95\textwidth,cols=9,pos=h]
  \caption{The test error of ResNeXt-29~(8x64d) with different settings on CIFAR-10.Here, `ep.' indicates epochs. `CA' and `CD' are cosine annealing and curriculum dropout. `D', `SD', and 	`SS' indicate standard dropout, SpatialDropout, and SelectScale respectively.}
  \label{resnext}
  \centering
  \begin{tabular}{p{12pt}p{12pt}p{12pt}p{12pt}p{12pt}p{12pt}p{12pt}p{12pt}p{12pt}}
    \toprule
   200 ep. & 600 ep. & CA & CD & no drop & D & SD & SS & error (\%)\\
     \midrule
    \checkmark &  &  &  & \checkmark &  &  &  & 4.19	\\
    \midrule
    \checkmark &  & \checkmark &  & \checkmark &  &  &  & 4.07	\\
    \midrule
     & \checkmark & \checkmark &  & \checkmark &  &  &  & 3.51	\\
     \midrule
     & \checkmark & \checkmark &  &  & \checkmark &  &  & 3.69	\\
     \midrule
     & \checkmark & \checkmark & \checkmark  &  & \checkmark &  &  & 3.66	\\
     \midrule
     & \checkmark & \checkmark & \checkmark  &  &  & \checkmark &  & 3.35	\\
     \midrule
     & \checkmark & \checkmark & \checkmark  &  &  &  & \checkmark & 3.24	\\
    
    \bottomrule
  \end{tabular}
\end{table}

%\newcommand{\tabincell}[2]{\begin{tabular}{@{}#1@{}}#2\end{tabular}}  
%\begin{table}[htp]
%  \caption{Test error of our method and competitive methods on CIFAR-10.}
%  \label{state-of-the-art}
%  \centering
%  \begin{tabular}{lcc}
%    \toprule
%    Method     & parameters(M) & error(\%) \\
%    \midrule
%    original-ResNet~\cite{he2016deep} & 10.2 & 7.93  \\
%    \midrule
%    stoc-depth-110~\cite{huang2016deep} & 1.7   & 5.23 \\
%    stoc-depth-1202 & 10.2 & 4.91 \\
%    \midrule
%    FractalNet~\cite{larsson2016fractalnet} & 38.6 & 5.22\\
%    with Dropout/Drop-path &38.6&4.60\\
%    \midrule
%    pre-ResNet~\cite{he2016identity} & 10.2 & 4.62   \\
%    \midrule
%    WRN-16-8~\cite{zagoruyko2016wide} & 36.5 & 4.00   \\ 
%    WRN~(Dropout) & 36.5 & 3.80     \\
%     \midrule
%    ResNeXt-29~(8x64d)~\cite{xie2016aggregated} & 34.4 & 3.65 \\
%    ResNeXt-29~(16x64d) & 68.1 & 3.58\\
%    \midrule
%    \multirow{2}{*}{\tabincell{l}{ResNeXt-29~(8x64d) \\SelectScale(ours)}}  
%     & \multirow{2}{*}{34.4}& \multirow{2}{*}{{\color{blue}3.24}} \\
%     \\
%    \midrule
%    \multirow{2}{*}{\tabincell{l}{DenseNet~\cite{huang2016densely}\\(L = 100,k = 24)}} & \multirow{2}{*}{27.2} & \multirow{2}{*}{3.74} \\
%    \\
%    
%    \multirow{2}{*}{\tabincell{l}{DenseNet-BC~\cite{huang2017densely} \\  (L = 190,k = 40)}} & \multirow{2}{*}{ 25.6 }& \multirow{2}{*}{3.46}\\    
%     \\
%    \midrule    
%    NASNet-A & 3.3 & 3.41 \\
%    \midrule   
%    
%    Shake-Shake-26 (2x96d) &26.2& {\color{red}2.86}\\
%    \bottomrule
%  \end{tabular}
%\end{table}

\subsection{Results on ImageNet}
ImageNet is a very large and complicated dataset and is very different from CIFAR. Many regularization methods that are useful in CIFAR can not improve the network performance on ImageNet~\cite{huang2016deep,devries2017improved}. In this section, we test our method on ImageNet. The tested network is ResNet-50. Drop methods are applied to the third and fourth groups after every BN~\cite{ioffe2015batch} layer and the drop rate in the third group is decreased by a factor following~\cite{ghiasi2018dropblock}. The factor is 0.5 in our experiments. The data retaining rate for dropout and SpatialDropout is 0.9. The top rate for SelectScale is 0.1 and the data retaining rate is 0.2. Curriculum drop is utilized.

%used before and after the $3\times3$ convolutional layers in residual blocks. For SclaeFilter, $q=0.4$ is used. The data retaining rate for DropFilter and standard dropout is 0.9.

As shown in Table~\ref{imagenet}, standard dropout increases the test error. SpatialDropout almost makes no difference on the network performance. Both these two methods introduce too much noise to the network. SelectScale is more effective in adjusting the learned patterns and decreases the top-1 and top-5 error by 0.5\% and 0.3\% respectively. 
%We conjecture that regularization on ImageNet is not as important as that on CIFAR. Regularization methods with fewer side effects (e.g. introducing noise) will perform better.

%SelectScale introduces less unnecessary noise during training.  that is why it can improve the network while 

\begin{figure}

\centering
\includegraphics[width=0.99\linewidth,height=0.55\linewidth]{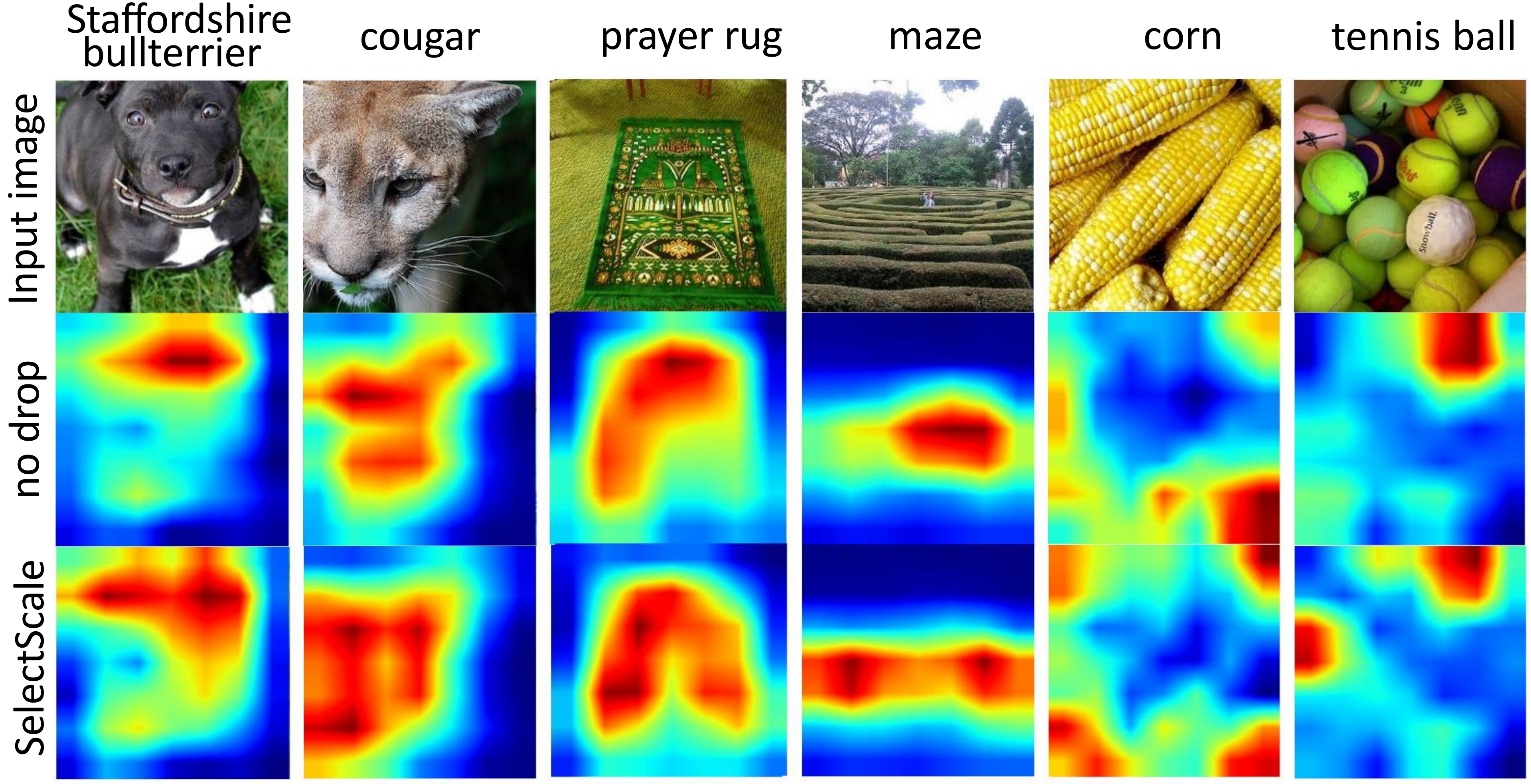} 

\caption{~The class activation maps for ResNet-50 on the validation set of ImageNet. These maps are generated by the $7\times7$ feature maps in group 4.}
\label{fig:cam-test}

\end{figure}

%ScaleFilter performs best and improve the performance by about 1.1\%. DropFilter and ScaleFilter can not only work well on tiny image datasets, but also remarkably improve the results of CNNs on bigger image dataset. 

%\newcommand{\tabincell}[2]{\begin{tabular}{@{}#1@{}}#2\end{tabular}}  
%
%\begin{table}[htp]
%  \caption{The top 5 test error(\%) of ResNet152 on the subset of ImageNet. The retaining rate for standard dropout and DropFilter is 0.9. ScaleFilter uses $q =0.4$ for this network.}
%  \label{imagenet}
%  \centering
%  \begin{tabular}{cccc}
%    \toprule
%    augmentation & method & top 1(\%) & top 5(\%)\\
%     \midrule
%    \multirow{4}{*}{random crop} & without data drop&	7.78 &24.98	\\
%   	\cline{2-4} 
%   & standard dropout&	7.46 &	24.60\\
%    \cline{2-4}
%     & DropFilter&	6.69 &	23.30\\
%    \cline{2-4}
%   & ScaleFilter & - &-\\
% \midrule
%  \multirow{4}{*}{ \tabincell{c}{random crop \\ + \\scale}} & without data drop&	- &-	\\
%   	\cline{2-4} 
%   & standard dropout&	- &	-\\
%    \cline{2-4}
%     & DropFilter&	- &-	\\
%    \cline{2-4}
%   & ScaleFilter & - &-\\
% 
%    \bottomrule
%  \end{tabular}
%\end{table}

\begin{table}[thbp]
  \caption{The test error of ResNet-50 with data drop regularization on ImageNet.}
  \label{imagenet}
  \centering
  \begin{tabular}{ccc}
    \toprule
    method &  top-1(\%) & top-5(\%)\\
     \midrule
    without data drop&	23.89 &	7.06\\
   	\midrule
    standard dropout&	23.97 & 7.12	\\
    \midrule
   SpatialDropout&	23.76 & 7.04\\
\midrule
   SelectScale&	23.38& 6.78	\\
    \bottomrule
  \end{tabular}
\end{table}

\subsection{The patterns learned by SelectScale}
We generate the class activation mapping (CAM) with the features before the global average pooling layer for original ResNet-50 and ResNet-50 with SelectScale. The shown examples in Figure~\ref{fig:cam-test} are selected from the validation set. As can be seen, SelectScale can find more patterns. During training, SelectScale continuously changes the learned patterns, which encourages the network to learn new patterns and prevent them from overfitting learned patterns. That is why networks using SelectScale are more robust and perform better.

\section{Discussion}
There are two hyperparameters that need to be tuned in SelectScale. The data retaining rate in SelectScale behaves like that in standard dropout. The network performance will increase as the retaining rate decreases before achieving the best retaining rate. After that, the performance gain will begin to decrease. The other hyperparameter is top rate. The bigger the top rate is, the more features will be changed. In our experiments, most tuned top rates fall in \{0.1, 0.2\}. 
%These top rates are big enough since dropping 6\% of the feature maps can disable a network as shown in Figure~\ref{fig:sdrop_cdrop}.
We suggest using these two rates if one does not want to tune the top rate.

\section{Conclusions}
In this paper, we present SelectScale to change the important learned features in networks effectively. By adjusting the learned patterns, the networks are forced to learn more patterns from images. Thus the networks with SelectScale become more robust and accurate. Compared to standard dropout and SpatialDropout which drop all features randomly, SelectScale is more effective and introduces less noise. 

The method in this paper is fast and parameter-free. It can provide stable improvement for different networks on different datasets. Additionally, our feature selection method does not rely on data drop methods. Given a new data drop method, it is easy to construct a ``selecting version'' according to our feature ranking strategy. We hope that SelectScale may become a new tool in helping researchers train deep neural networks.

%Selectout and SelectScale can also be understood by regularization. Selectout and SelectScale provide stronger regularization than their counterparts with the same data retaining rate. In other words, they are more modest in changing the whole feature distribution compared to other data drop regularization methods. That is why SelectScale consistently outperforms other methods for different networks and datasets.

\section*{Acknowledgements}
This work is supported by the National Key R\&D Program of China under Grant No. 2017YFB1301100, the National Natural Science Foundation of China under Grant Nos. 61976012 and 61772060, and CERNET Innovation Project under Grant No. NGII20170315.

%This work was supported by the National Natural Science Foundation of China ( 61572060, U1536107, 61472024, 61602024) and the CERNET Innovation Project (NGII20151004,  NGII20160316).

%In this paper, we propose DropFilter which is a new regularization method for convolutions. Unlike standard dropout, DropFilter only aims to reduce the co-adaptations inter filters. ScaleFilter randomly scales the outputs of filters. CNNs can be regularized by ScaleFilter when DropFilter is too coarse for the networks. Using the methods in this paper, we significantly improve the performance of different architecture CNNs on CIFAR and the subset of ImageNet. In addition, DropFilter nearly does not introduce extra computation like standard dropout.
%Multi-path networks have achieved state-of-the-art results~\cite{gastaldi2017shake,zoph2017learning} on CIFAR and ImageNet with DropPath, which demonstrates that data drop regularization methods are promising for improving deep neural networks. We bridge the gaps not only between multi-path networks and single path networks but also between standard dropout and DropPath. We hope that more concerns could be paid to data drop regularization methods. We believe that this is a promising perspective to understand and improve deep neural networks.

\bibliographystyle{named}
\bibliography{ijcai20}

\end{document}